\documentclass[10pt,twocolumn,letterpaper]{article}

\usepackage{cvpr}
\usepackage{times}
\usepackage{epsfig}
\usepackage{graphicx}

\usepackage{amsmath,amssymb,amsfonts,dsfont,bm,mathrsfs}
\usepackage{caption}
\usepackage{xcolor}
\usepackage{booktabs,multirow,adjustbox}
\usepackage{float}
\usepackage[pagebackref=true,breaklinks=true,letterpaper=true,colorlinks,bookmarks=false]{hyperref}
\newcommand{\R}{\mathbb{R}}      
\newcommand{\E}{\mathbb{E}}      
\newcommand{\Prob}{{\rm P}}      
\newcommand{\Norm}{\mathcal{N}}  
\newcommand{\0}{{\rm\bf 0}}      
\newcommand{\I}{{\rm\bf I}}      
\newcommand{\z}{{\rm\bf z}}      
\newcommand{\Z}{\mathcal{Z}}     
\newcommand{\w}{{\rm\bf w}}      
\newcommand{\W}{\mathcal{W}}     
\newcommand{\x}{{\rm\bf x}}      
\newcommand{\X}{\mathcal{X}}     
\newcommand{\s}{{\rm\bf s}}      
\renewcommand{\S}{\mathcal{S}}   
\newcommand{\n}{{\rm\bf n}}      
\newcommand{\N}{{\rm\bf N}}      
\newcommand{\dis}{{\rm d}}       

\def\cvprPaperID{5891}
\cvprfinalcopy

\begin{document}

\title{Interpreting the Latent Space of GANs for Semantic Face Editing}
\author{
  Yujun Shen\textsuperscript{1},
  Jinjin Gu\textsuperscript{2},
  Xiaoou Tang\textsuperscript{1},
  Bolei Zhou\textsuperscript{1} \\
  \textsuperscript{1}The Chinese University of Hong Kong \quad
  \textsuperscript{2}The Chinese University of Hong Kong, Shenzhen \\
  {\tt\small
    \{sy116, xtang, bzhou\}@ie.cuhk.edu.hk,
    jinjingu@link.cuhk.edu.cn
  }
}

\twocolumn[{
\renewcommand\twocolumn[1][]{#1}
\maketitle
\begin{center}
  \includegraphics[width=0.95\linewidth]{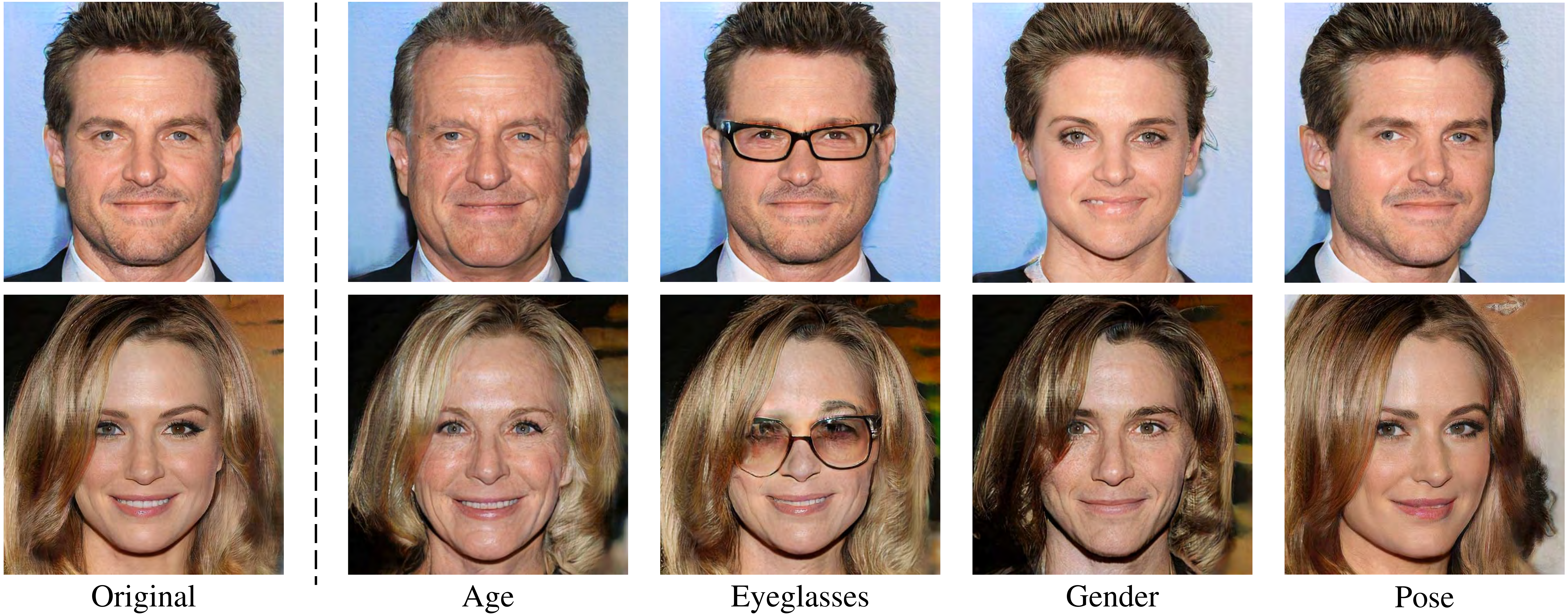}
  \vspace{-7pt}
  \captionsetup{type=figure,font=small}
  \caption{
    Manipulating various facial attributes through varying the latent codes of a well-trained GAN model.
    The first column shows the original synthesis from PGGAN \cite{pggan}, while each of the other columns shows the results of manipulating a specific attribute.
  }
  \label{fig:teaser}
  \vspace{5pt}
\end{center}
}]


\begin{abstract}
Despite the recent advance of Generative Adversarial Networks (GANs) in high-fidelity image synthesis, there lacks enough understanding of how GANs are able to map a latent code sampled from a random distribution to a photo-realistic image.
Previous work assumes the latent space learned by GANs follows a distributed representation but observes the vector arithmetic phenomenon.
In this work, we propose a novel framework, called InterFaceGAN, for semantic face editing by interpreting the latent semantics learned by GANs.
In this framework, we conduct a detailed study on how different semantics are encoded in the latent space of GANs for face synthesis.
We find that the latent code of well-trained generative models actually learns a disentangled representation after linear transformations.
We explore the disentanglement between various semantics and manage to decouple some entangled semantics with subspace projection, leading to more precise control of facial attributes.
Besides manipulating gender, age, expression, and the presence of eyeglasses, we can even vary the face pose as well as fix the artifacts accidentally generated by GAN models.
The proposed method is further applied to achieve real image manipulation when combined with GAN inversion methods or some encoder-involved models.
Extensive results suggest that learning to synthesize faces spontaneously brings a disentangled and controllable facial attribute representation.\footnote{Code and models are available at \href{https://genforce.github.io/interfacegan/}{this link}.}
\end{abstract}

\section{Introduction}\label{sec:introduction}
Generative Adversarial Networks (GANs) \cite{gan} have significantly advanced image synthesis in recent years.
The rationale behind GANs is to learn the mapping from a latent distribution to the real data through adversarial training.
After learning such a non-linear mapping, GAN is capable of producing photo-realistic images from randomly sampled latent codes.
However, it is uncertain how semantics originate and are organized in the latent space.
Taking face synthesis as an example, when sampling a latent code to produce an image, how the code is able to determine various semantic attributes (\emph{e.g.}, gender and age) of the output face, and how these attributes are entangled with each other?

Existing work typically focuses on improving the synthesis quality of GANs \cite{sagan,sngan,pggan,biggan,stylegan}, however, few efforts have been made on studying what a GAN actually learns with respect to the latent space.
Radford \emph{et al.} \cite{dcgan} first observes the vector arithmetic property in the latent space.
A recent work \cite{gan_dissection} further shows that some units from intermediate layers of the GAN generator are specialized to synthesize certain visual concepts, such as sofa and TV for living room generation.
Even so, there lacks enough understanding of how GAN connects the latent space and the image semantic space, as well as how the latent code can be used for image editing.

In this paper, we propose a framework \emph{InterFaceGAN}, short for \emph{Inter}preting \emph{Face GAN}s, to identify the semantics encoded in the latent space of well-trained face synthesis models and then utilize them for semantic face editing.
Beyond the vector arithmetic property, this framework provides both theoretical analysis and experimental results to verify that \emph{linear} subspaces align with different \emph{true-or-false} semantics emerging in the latent space.
We further study the disentanglement between different semantics and show that we can decouple some entangled attributes (\emph{e.g.}, old people are more likely to wear eyeglasses then young people) through the linear subspace projection.
These disentangled semantics enable precise control of facial attributes with any given GAN model \emph{without retraining}.

Our contributions are summarized as follows:
\begin{itemize}
  \vspace{-8pt}
  \setlength{\itemsep}{0pt}
  \setlength{\parsep}{0pt}
  \setlength{\parskip}{0pt}
  \item We propose InterFaceGAN to explore how a single or multiple semantics are encoded in the latent space of GANs, such as PGGAN \cite{pggan} and StyleGAN \cite{stylegan}, and observe that GANs spontaneously learn various latent subspaces corresponding to specific attributes. These attribute representations become disentangled after some linear transformations.
  \item We show that InterFaceGAN enables semantic face editing with any \emph{fixed} pre-trained GAN model. Some results are shown in Fig.\ref{fig:teaser}. Besides gender, age, expression, and the presence of eyeglasses, we can noticeably also vary the face pose or correct some artifacts produced by GANs.
  \item We extend InterFaceGAN to real image editing with GAN inversion methods and encoder-involved models. We successfully manipulate the attributes of real faces by simply varying the latent code, even with GANs that are not specifically designed for the editing task.
\end{itemize}

\subsection{Related Work}\label{subsec:related-work}

\noindent\textbf{Generative Adversarial Networks.}
GAN \cite{gan} has brought wide attention in recent years due to its great potential in producing photo-realistic images \cite{wgan,wgan_gp,began,sagan,sngan,pggan,biggan,stylegan}.
It typically takes a sampled latent code as the input and outputs an image synthesis.
To make GANs applicable for real image processing, existing methods proposed to reverse the mapping from the latent space to the image space \cite{icgan,zhu2016generative,invertibility,bau2019seeing,gu2020image} or learn an additional encoder associated with the GAN training \cite{ali,bigan,lia}.
Despite this tremendous success, little work has been done on understanding how GANs learn to connect the input latent space with the semantics in the real visual world.

\noindent\textbf{Study on Latent Space of GANs.}
Latent space of GANs is generally treated as Riemannian manifold \cite{gan_metrics,latent_oddity,kuhnel2018latent}.
Prior work focused on exploring how to make the output image vary smoothly from one synthesis to another through interpolation in the latent space, regardless of whether the image is semantically controllable \cite{feature_based_metrics,riemannian_geometry}.
GLO \cite{glo} optimized the generator and latent code simultaneously to learn a better latent space.
However, the study on how a well-trained GAN is able to encode different semantics inside the latent space is still missing.
Some work has observed the vector arithmetic property \cite{dcgan,feature_interpolation}.
Beyond that, this work provides a detailed analysis of the semantics encoded in the latent space from both the property of a single semantic and the disentanglement of multiple semantics.
Some concurrent work also explores the latent semantics learned by GANs.
Jahanian \emph{et al.} \cite{gansteerability} studies the steerability of GANs concerning camera motion and image color tone.
Goetschalckx \emph{et al.} \cite{goetschalckx2019ganalyze} improves the memorability of the output image.
Yang \emph{et al.} \cite{yang2019semantic} explores the hierarchical semantics in the deep generative representations for scene synthesis.
Unlike them, we focus on facial attributes emerging in GANs for face synthesis and extend our method to real image manipulation.

\noindent\textbf{Semantic Face Editing with GANs.}
Semantic face editing aims at manipulating facial attributes of a given image.
Compared to unconditional GANs which can generate image arbitrarily, semantic editing expects the model to only change the target attribute but maintain other information of the input face.
To achieve this goal, current methods required carefully designed loss functions \cite{acgan,infogan,drgan}, introduction of additional attribute labels or features \cite{fader,ffgan,opensetgan,elegant,facefeatgan}, or special architectures \cite{sdgan,faceidgan} to train new models.
However, the synthesis resolution and quality of these models are far behind those of native GANs, like PGGAN \cite{pggan} and StyleGAN \cite{stylegan}.
Different from previous learning-based methods, this work explores the interpretable semantics inside the latent space of \emph{fixed} GAN models, and \emph{turns unconstrained GANs to controllable GANs} by varying the latent code.

\section{Framework of InterFaceGAN}\label{sec:interfacegan}
In this section, we introduce the framework of InterFaceGAN, which first provides a rigorous analysis of the semantic attributes emerging in the latent space of well-trained GAN models, and then constructs a manipulation pipeline of leveraging the semantics in the latent code for facial attribute editing.

\subsection{Semantics in the Latent Space}\label{subsec:semantics-interpretation}
Given a well-trained GAN model, the generator can be formulated as a deterministic function $g: \Z\rightarrow\X$.
Here, $\Z\subseteq\R^{d}$ denotes the $d$-dimensional latent space, for which Gaussian distribution $\Norm(\0, \I_d)$ is commonly used \cite{sngan,pggan,biggan,stylegan}.
$\X$ stands for the image space, where each sample $\x$ possesses certain semantic information, like gender and age for face model.
Suppose we have a semantic scoring function $f_S: \X\rightarrow\S$, where $\S\subseteq\R^m$ represents the semantic space with $m$ semantics.
We can bridge the latent space $\Z$ and the semantic space $\S$ with $\s = f_S(g(\z))$, where $\s$ and $\z$ denote the semantic scores and the sampled latent code respectively.

\vspace{2pt}\noindent\textbf{Single Semantic.}
It has been widely observed that when linearly interpolating two latent codes $\z_1$ and $\z_2$, the appearance of the corresponding synthesis changes continuously \cite{dcgan,biggan,stylegan}.
It implicitly means that the semantics contained in the image also change gradually.
According to \emph{Property 1}, the linear interpolation between $\z_1$ and $\z_2$ forms a direction in $\Z$, which further defines a hyperplane.
We therefore make an assumption\footnote{This assumption is empirically verified in Sec.\ref{subsec:latent-space-separation}.} that for any binary semantic (\emph{e.g.}, male \emph{v.s.} female), there exists a hyperplane in the latent space serving as the separation boundary.
Semantic remains the same when the latent code walks within the same side of the hyperplane yet turns into the opposite when across the boundary.

Given a hyperplane with a unit normal vector $\n\in\R^d$, we define the ``distance'' from a sample $\z$ to this hyperplane as
\begin{align}
  \dis(\n, \z)= \n^T\z.  \label{eq:distance}
\end{align}
Here, $\dis(\cdot,\cdot)$ is not a strictly defined distance since it can be negative.
When $\z$ lies near the boundary and is moved toward and across the hyperplane, both the ``distance'' and the semantic score vary accordingly.
And it is just at the time when the ``distance'' changes its numerical sign that the semantic attribute reverses.
We therefore expect these two to be linearly dependent with
\begin{align}
  f(g(\z)) = \lambda\dis(\n,\z),  \label{eq:linear-dependency}
\end{align}
where $f(\cdot)$ is the scoring function for a particular semantic, and $\lambda > 0$ is a scalar to measure how fast the semantic varies along with the change of distance.
According to \emph{Property 2}, random samples drawn from $\Norm(\0,\I_d)$ are very likely to locate close enough to a given hyperplane.
Therefore, the corresponding semantic can be modeled by the linear subspace that is defined by $\n$.

\vspace{2pt}\emph{\textbf{Property 1}
Given $\n\in\R^d$ with $\n\neq\0$, the set $\{\z\in\R^d:\n^T\z=0\}$ defines a hyperplane in $\R^d$, and $\n$ is called the normal vector. All vectors $\z\in\R^d$ satisfying $\n^T\z>0$ locate from the same side of the hyperplane.
}

\vspace{2pt}\emph{\textbf{Property 2}
Given $\n\in\R^d$ with $\n^T\n=1$, which defines a hyperplane, and a multivariate random variable $\z\sim\Norm(\0,\I_d)$, we have $\Prob(|\n^T\z|\leq2\alpha~\sqrt{\frac{d}{d-2}})\geq(1-3e^{-c d})(1-\frac{2}{\alpha}e^{-\alpha^2/2})$ for any $\alpha\geq1$ and $d\geq4$. Here, $\Prob(\cdot)$ stands for probability and $c$ is a fixed positive constant.\footnote{When $d=512$, we have $P(|\n^T\z|>5.0)<1e^{-6}$. It suggests that almost all sampled latent codes are expected to locate within 5 unit-length to the boundary. Proof can be found in \textbf{Appendix}.}
}

\vspace{2pt}\noindent\textbf{Multiple Semantics.}
When the case comes to $m$ different semantics, we have
\begin{align}
  \s \equiv f_S(g(\z))=\Lambda\N^T\z,  \label{eq:multiple-semantics}
\end{align}
where $\s = [s_1, \dots, s_m]^T$ denotes the semantic scores, $\Lambda = \text{diag}(\lambda_1, \dots, \lambda_m)$ is a diagonal matrix containing the linear coefficients, and $\N = [\n_1, \dots, \n_m]$ indicates the separation boundaries.
Aware of the distribution of random sample $\z$, which is $\Norm(\0,\I_d)$, we can easily compute the mean and covariance matrix of the semantic scores $\s$ as
\begin{align}
  \bm{\mu}_\s &= \E(\Lambda\N^T\z) = \Lambda\N^T\E(\z) = \0,  \label{eq:score-mean}  \\
  \bm{\Sigma}_\s &= \E(\Lambda\N^T\z\z^T\N\Lambda^T) = \Lambda\N^T\E(\z\z^T)\N\Lambda^T  \notag \\
                 &= \Lambda\N^T\N\Lambda.  \label{eq:score-cov}
\end{align}

We therefore have $\s\sim\Norm(\0,\bm{\Sigma}_\s)$, which is a multivariate normal distribution.
Different entries of $\s$ are disentangled if and only if $\bm{\Sigma}_\s$ is a diagonal matrix, which requires $\{\n_1, \dots, \n_m\}$ to be orthogonal with each other.
If this condition does not hold, some semantics will correlate with each other and $\n_i^T\n_j$ can be used to measure the entanglement between the $i$-th and $j$-th semantics.

\begin{figure}[t]
  \centering
  \includegraphics[width=0.75\linewidth]{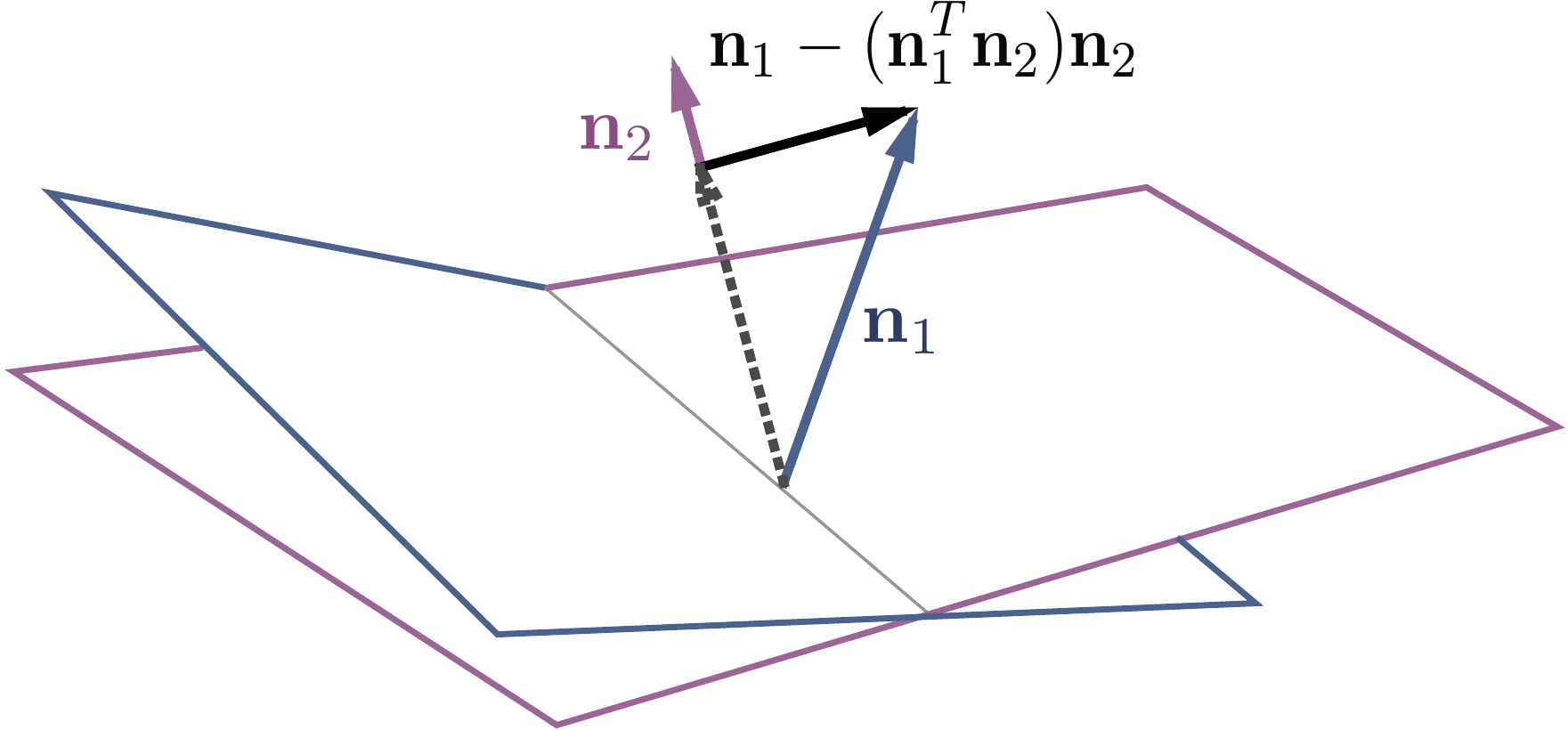}
  \vspace{-8pt}
  \captionsetup{font=small}
  \caption{
    Illustration of the conditional manipulation in subspace.
    The projection of $\n_1$ onto $\n_2$ is subtracted from $\n_1$, resulting in a new direction $\n_1 - (\n_1^T\n_2)\n_2$.
  }
  \label{fig:subspace}
  \vspace{-10pt}
\end{figure}

\subsection{Manipulation in the Latent Space}\label{subsec:semantics-manipulation}
In this part, we introduce how to use the semantics found in latent space for image editing.

\vspace{2pt}\noindent\textbf{Single Attribute Manipulation.}
According to Eq.\eqref{eq:linear-dependency}, to manipulate the attribute of a synthesized image, we can easily edit the original latent code $\z$ with $\z_{edit}=\z + \alpha \n$.
It will make the synthesis look more positive on such semantic with $\alpha>0$, since the score becomes $f(g(\z_{edit}))=f(g(\z)) + \lambda\alpha$ after editing.
Similarly, $\alpha < 0$ will make the synthesis look more negative.

\vspace{2pt}\noindent\textbf{Conditional Manipulation.}
When there is more than one attribute, editing one may affect another since some semantics can be coupled with each other.
To achieve more precise control, we propose \emph{conditional manipulation} by manually forcing $\N^T\N$ in Eq.\eqref{eq:score-cov} to be diagonal.
In particular, we use projection to orthogonalize different vectors.
As shown in Fig.\ref{fig:subspace}, given two hyperplanes with normal vectors $\n_1$ and $\n_2$, we find a projected direction $\n_1 - (\n_1^T\n_2)\n_2$, such that moving samples along this new direction can change ``attribute 1'' without affecting ``attribute 2''.
We call this operation as conditional manipulation.
If there is more than one attribute to be conditioned on, we just subtract the projection from the primal direction onto the plane that is constructed by all conditioned directions.

\vspace{2pt}\noindent\textbf{Real Image Manipulation.}
Since our approach enables semantic editing from the latent space of a \emph{fixed} GAN model, we need to first map a real image to a latent code before performing manipulation.
For this purpose, existing methods have proposed to directly optimize the latent code to minimize the reconstruction loss \cite{invertibility}, or to learn an extra encoder to invert the target image back to latent space \cite{zhu2016generative,bau2019seeing}.
There are also some models that have already involved an encoder along with the training process of GANs \cite{ali,bigan,lia}, which we can directly use for inference.

\section{Experiments}\label{sec:experiments}
In this section, we evaluate InterFaceGAN with state-of-the-art GAN models, PGGAN \cite{pggan} and StyleGAN \cite{stylegan}.
Specifically, the experiments in Sec.\ref{subsec:latent-space-separation}, Sec.\ref{subsec:latent-space-manipulation}, and Sec.\ref{subsec:conditional-manipulation} are conducted on PGGAN to interpret the latent space of the traditional generator.
Experiments in Sec.\ref{subsec:results-on-stylegan} are carried out on StyleGAN to investigate the style-based generator and also compare the differences between the two sets of latent representations in StyleGAN.
We also apply our approach to real images in Sec.\ref{subsec:real-image-manipulation} to see how the semantics implicitly learned by GANs can be applied to real face editing.
Implementation details can be found in \textbf{Appendix}.

\subsection{Latent Space Separation}\label{subsec:latent-space-separation}
As mentioned in Sec.\ref{subsec:semantics-interpretation}, our framework is based on an assumption that for any binary attribute, there exists a hyperplane in latent space such that all samples from the same side are with the same attribute.
Accordingly, we would like to first evaluate the correctness of this assumption to make the remaining analysis considerable.

We train five independent linear SVMs on pose, smile, age, gender, eyeglasses, and then evaluate them on the validation set ($6K$ samples with high confidence level on attribute scores) as well as the entire set ($480K$ random samples).
Tab.\ref{tab:separation} shows the results.
We find that all linear boundaries achieve over 95\% accuracy on the validation set and over 75\% on the entire set, suggesting that for a binary attribute, there exists a linear hyperplane in the latent space that can well separate the data into two groups.

\begin{figure}[t]
  \centering
  \includegraphics[width=0.9\linewidth]{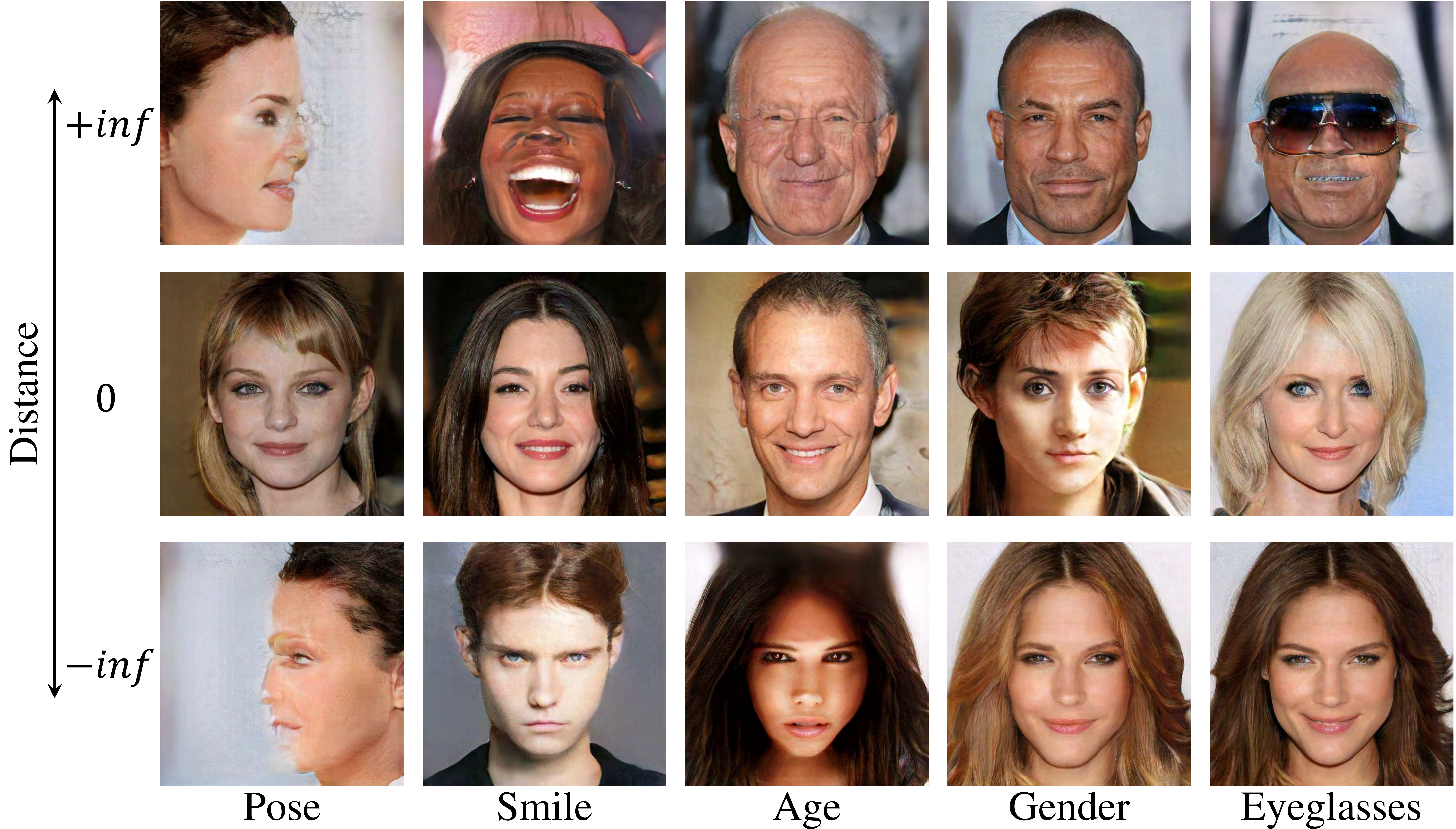}
  \vspace{-10pt}
  \captionsetup{font=small}
  \caption{
    Synthesis samples with the distance near to (middle row) and extremely far away from (top and bottom rows) the separation boundary.
    Each column corresponds to a particular attribute.
  }
  \label{fig:separation}
  \vspace{-8pt}
\end{figure}

We also visualize some samples in Fig.\ref{fig:separation} by ranking them with the distance to the decision boundary.
Note that those extreme cases (first and last row in Fig.\ref{fig:separation}) are very unlikely to be directly sampled, instead constructed by moving a latent code towards the normal direction ``infinitely''.
From Fig.\ref{fig:separation}, we can tell that the positive samples and negative samples are distinguishable to each other with respect to the corresponding attribute.

\setlength{\tabcolsep}{6pt}
\begin{table}[t]
  \captionsetup{font=small}
  \caption{
    Classification accuracy (\%) on separation boundaries in latent space with respect to different attributes.
  }
  \vspace{-8pt}
  \label{tab:separation}
  \centering\small
  \begin{tabular}{c|c|c|c|c|c}
    \hline
    Dataset    &  Pose & Smile &  Age & Gender & Eyeglasses \\ \hline
    Validation & 100.0 &  96.9 & 97.9 &   98.7 &       95.6 \\ \hline
    All        &  90.3 &  78.5 & 75.3 &   84.2 &       80.1 \\ \hline
  \end{tabular}
  \vspace{-10pt}
\end{table}

\subsection{Latent Space Manipulation}\label{subsec:latent-space-manipulation}
In this part, we verify whether the semantics found by InterFaceGAN are manipulable.

\begin{figure*}[t]
  \centering
  \includegraphics[width=0.95\linewidth]{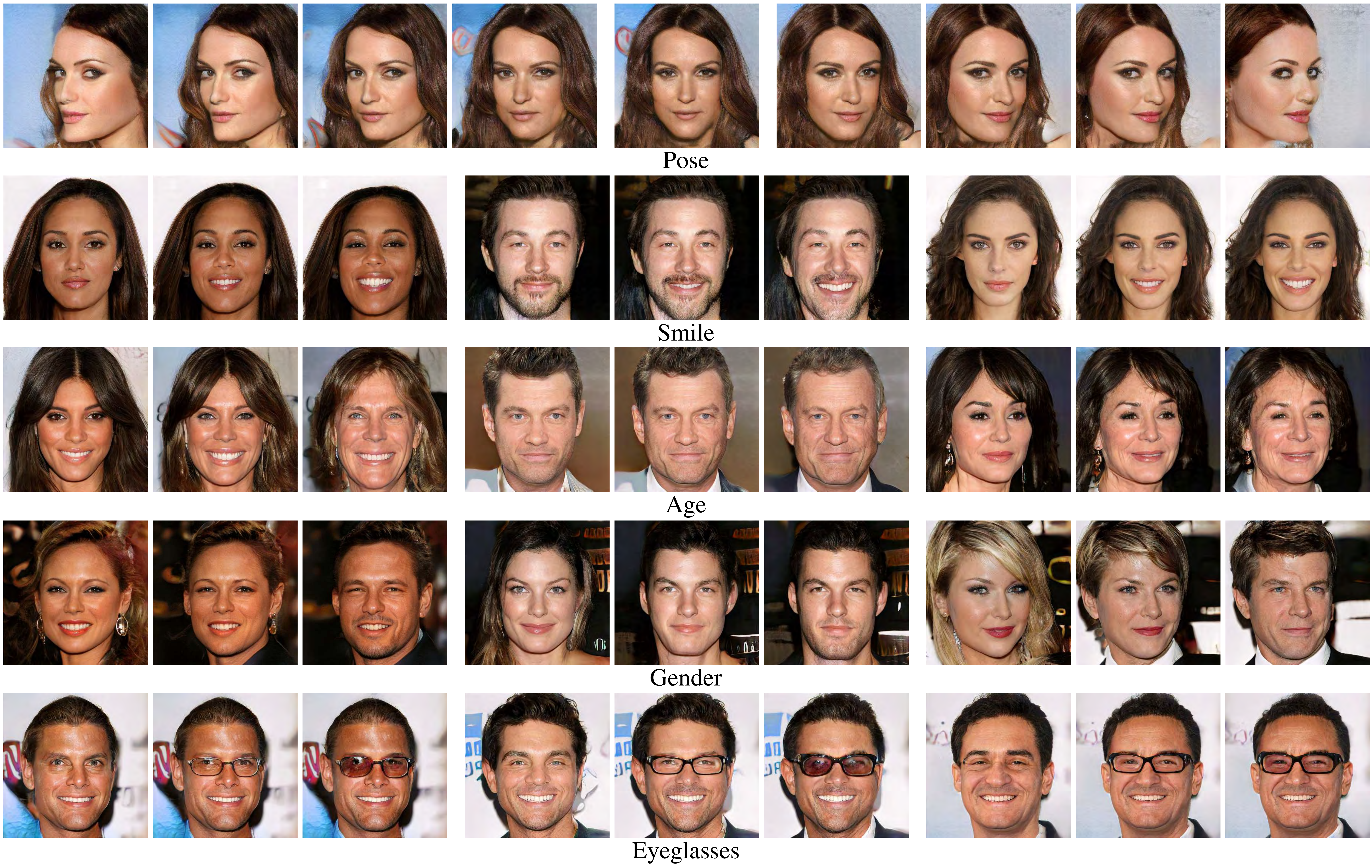}
  \vspace{-10pt}
  \captionsetup{font=small}
  \caption{
    Single attribute manipulation results.
    The first row shows the same person under gradually changed poses.
    The following rows correspond to the results of manipulating four different attributes.
    For each set of three samples in a row, the central one is the original synthesis, while the left and right stand for the results by moving the latent code along negative and positive direction respectively.
  }
  \label{fig:manipulation}
  \vspace{-5pt}
\end{figure*}

\begin{figure*}[t]
  \centering
  \includegraphics[width=0.95\linewidth]{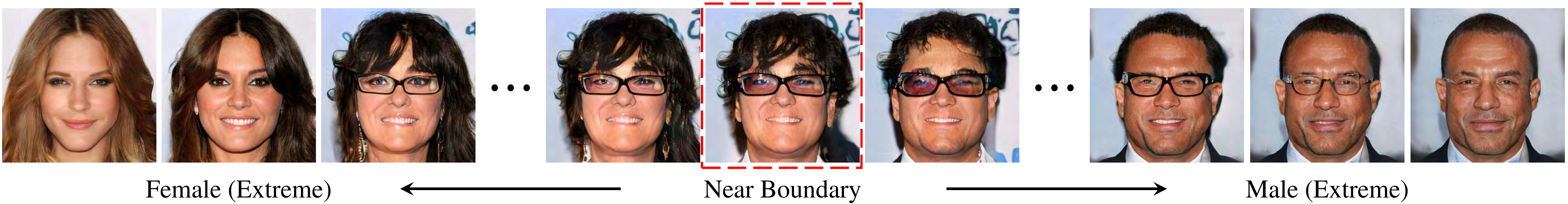}
  \vspace{-10pt}
  \captionsetup{font=small}
  \caption{
    Illustration of the distance effect by taking gender manipulation as an example.
    The image in the red dashed box stands for the original synthesis.
    Our approach performs well when the latent code locates close to the boundary.
    However, when the distance keeps increasing, the synthesized images are no longer like the same person.
  }
  \label{fig:limitation}
  \vspace{-10pt}
\end{figure*}

\vspace{2pt}\noindent\textbf{Manipulating Single Attribute.}
Fig.\ref{fig:manipulation} plots the manipulation results on five different attributes.
It suggests that our manipulation approach performs well on all attributes in both positive and negative directions.
Particularly on \emph{pose} attribute, we observe that even the boundary is searched by solving a bi-classification problem, moving the latent code can produce continuous changing.
Furthermore, although there lacks enough data with extreme poses in the training set, GAN is capable of imagining how profile faces should look like.
The same situation also happens on eyeglasses attribute.
We can manually create a lot of faces wearing eyeglasses despite the inadequate data in the training set.
These two observations provide strong evidence that GAN does not produce images randomly, but learns some interpretable semantics from the latent space.

\vspace{2pt}\noindent\textbf{Distance Effect of Semantic Subspace.}
When manipulating the latent code, we observe an interesting distance effect that the samples will suffer from severe changes in appearance if being moved too far from the boundary, and finally tend to become the extreme cases shown in Fig.\ref{fig:separation}.
Fig.\ref{fig:limitation} illustrates this phenomenon by taking gender editing as an instance.
Near-boundary manipulation works well.
When samples go beyond a certain region\footnote{We choose 5.0 as the threshold.}, however, the editing results are no longer like the original face anymore.
But this effect does not affect our understanding of the disentangled semantics in latent space.
That is because such extreme samples are very unlikely to be directly drawn from a standard normal distribution, which is pointed out in \textbf{Property 2} in Sec.\ref{subsec:semantics-interpretation}.
Instead, they are constructed manually by keeping moving a normally sampled latent code along a certain direction.
In this way, we can get a better interpretation on the latent semantics of GANs.

\begin{figure}[t]
  \centering
  \includegraphics[width=0.9\linewidth]{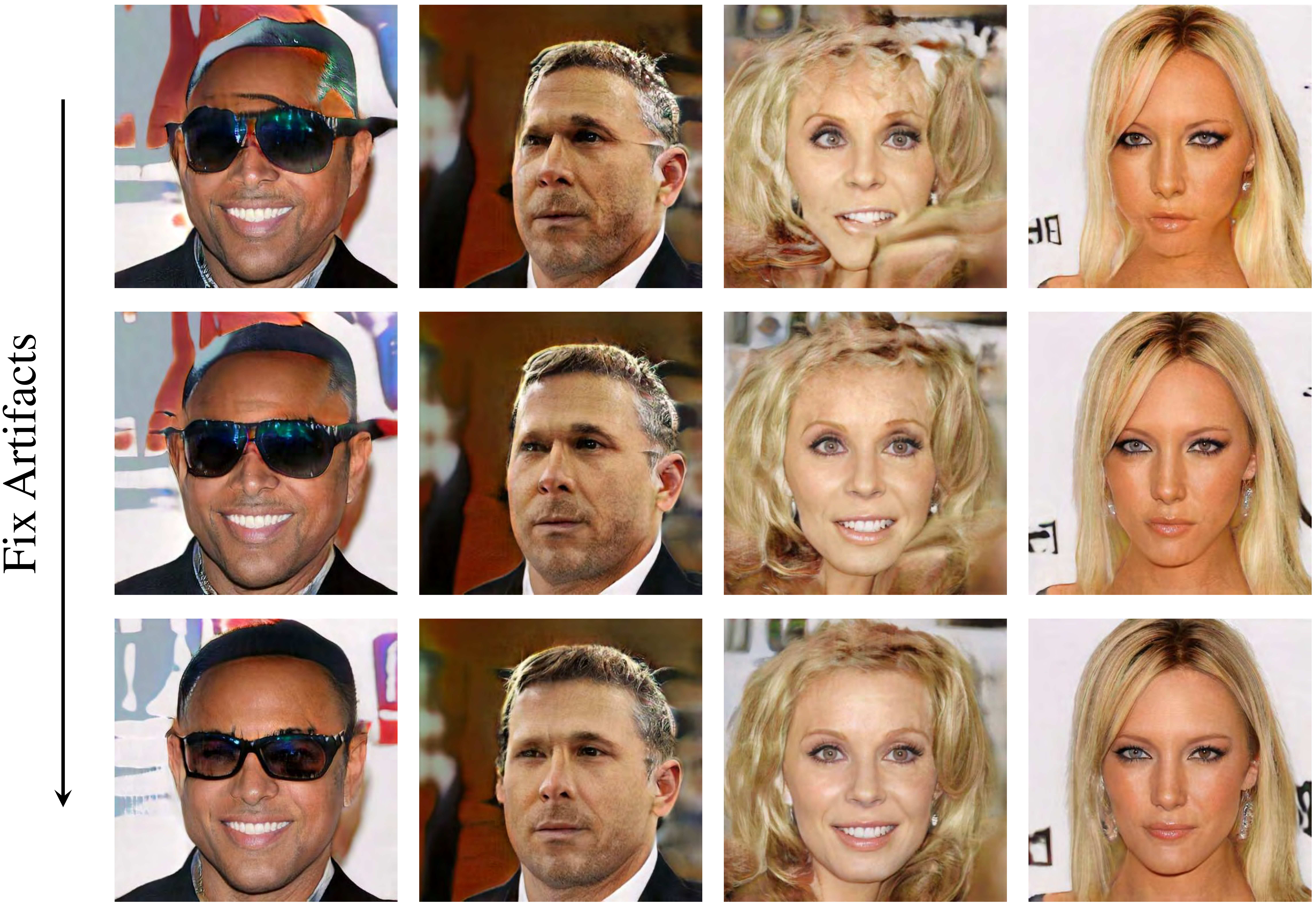}
  \vspace{-10pt}
  \captionsetup{font=small}
  \caption{
    Examples on fixing the artifacts that GAN has generated.
    First row shows some bad generation results, while the following two rows present the gradually corrected synthesis by moving the latent codes along the positive ``quality'' direction.
  }
  \label{fig:correction}
  \vspace{-30pt}
\end{figure}

\begin{figure*}[t]
  \centering
  \includegraphics[width=0.95\linewidth]{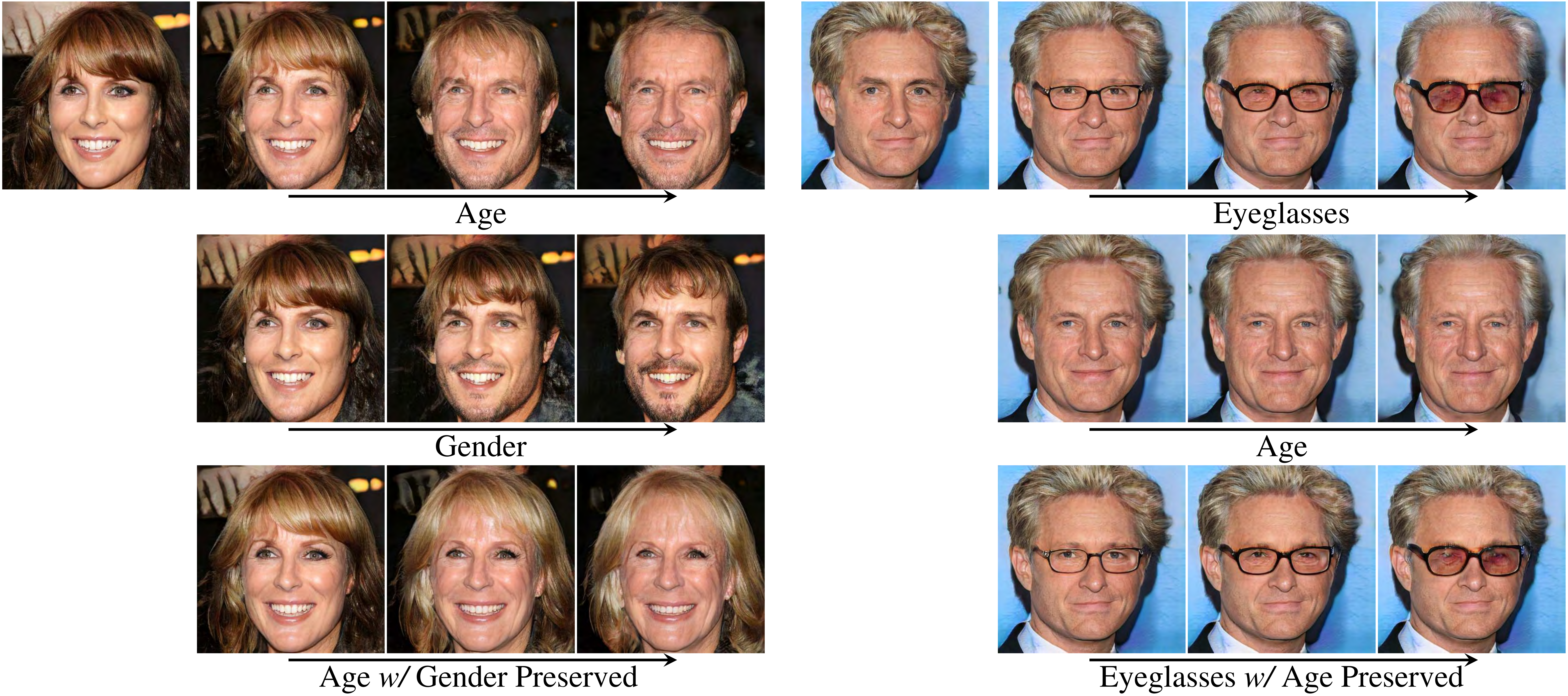}
  \vspace{-8pt}
  \captionsetup{font=small}
  \caption{
    Examples for conditional manipulation.
    The first two rows show the manipulation results along with the original directions learned by SVMs for two attributes independently.
    The last row edits the faces by varying one attribute with the other one unchanged.
  }
  \label{fig:condition}
  \vspace{-10pt}
\end{figure*}

\vspace{2pt}\noindent\textbf{Artifacts Correction.}
We further apply our approach to fix the artifacts that sometimes occurred in the synthesized outputs.
We manually labeled $4K$ bad synthesis and then trained a linear SVM to find the separation hyperplane, same as other attributes.
We surprisingly find that GAN also encodes such information in latent space.
Based on this discovery, we are capable of correcting some mistakes GAN has made in the generation process, as shown in Fig.\ref{fig:correction}.

\subsection{Conditional Manipulation}\label{subsec:conditional-manipulation}
In this section, we study the disentanglement between different attributes and evaluate the conditional manipulation approach.

\vspace{2pt}\noindent\textbf{Correlation between Attributes.}
Different from \cite{stylegan} which introduced perceptual path length and linear separability to measure the disentanglement property of latent space, we focus more on the relationships between different hidden semantics and study how they are coupled with each other.
Here, two different metrics are used to measure the correlation between two attributes.
(i) We compute the cosine similarity between two directions as $\cos(\n_1, \n_2)=\n_1^T \n_2$, where $\n_1$ and $\n_2$ stand for unit vectors.
(ii) We treat each attribute score as a random variable, and use the attribute distribution observed from all $500K$ synthesized data to compute the correlation coefficient $\rho$.
Here, we have $\rho_{A_1A_2}=\frac{Cov(A_1,A_2)}{\sigma_{A_1}\sigma_{A_2}}$, where $A_1$ and $A_2$ represent two random variables with respect to two attributes. $Cov(\cdot,\cdot)$ stands for covariance, and $\sigma$ denotes standard deviation.

\begin{figure}[t]
  \centering
  \includegraphics[width=0.95\linewidth]{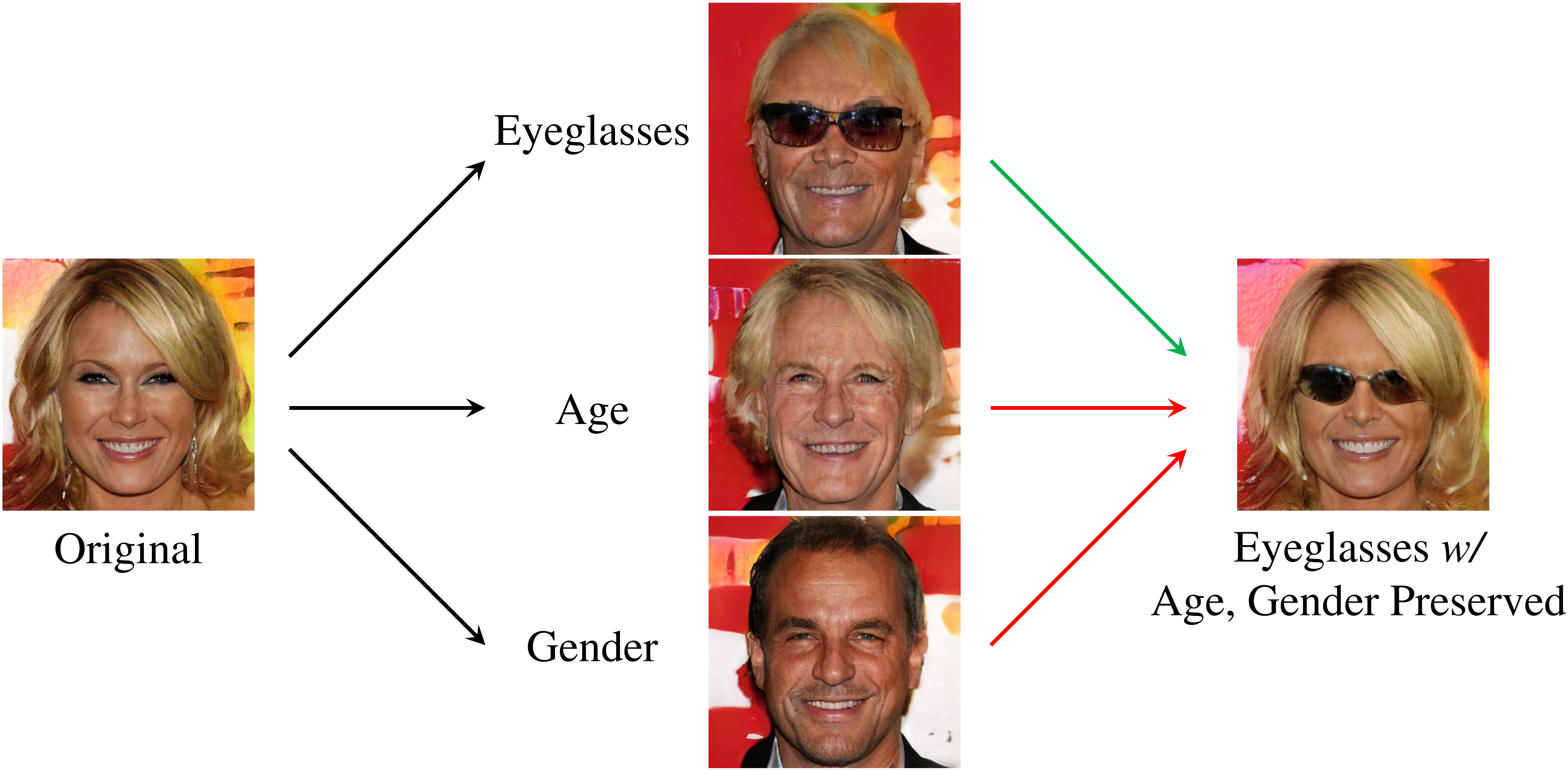}
  \vspace{-8pt}
  \captionsetup{font=small}
  \caption{
    Examples for conditional manipulation with more than one conditions.
    Left: Original synthesis. Middle: Manipulations along single boundary. Right: Conditional manipulation.
    \textbf{\textcolor{green}{Green}} arrow: Primal direction.
    \textbf{\textcolor{red}{Red}} arrows: Projection subtraction.
  }
  \label{fig:multiple-conditions}
  \vspace{-10pt}
\end{figure}

\setlength{\tabcolsep}{6pt}
\begin{table}[t]
  \captionsetup{font=small}
  \caption{
    Correlation matrix of attribute boundaries.
  }
  \label{tab:boundary-correlation}
  \vspace{-8pt}
  \centering\small
  \begin{tabular}{|c|c|c|c|c|c|}
    \hline
               & Pose & Smile &   Age & Gender & Eyeglasses \\ \hline
    Pose       & 1.00 & -0.04 & -0.06 &  -0.05 &      -0.04 \\ \hline
    Smile      &    - &  1.00 &  0.04 &  -0.10 &      -0.05 \\ \hline
    Age        &    - &     - &  1.00 &   0.49 &       0.38 \\ \hline
    Gender     &    - &     - &     - &   1.00 &       0.52 \\ \hline
    Eyeglasses &    - &     - &     - &      - &       1.00 \\ \hline
  \end{tabular}
  \vspace{-10pt}
\end{table}

\setlength{\tabcolsep}{6pt}
\begin{table}[t]
  \captionsetup{font=small}
  \caption{
    Correlation matrix of synthesized attribute distributions.
  }
  \label{tab:attribute-correlation}
  \vspace{-8pt}
  \centering\small
  \begin{tabular}{|c|c|c|c|c|c|}
    \hline
               & Pose & Smile &   Age & Gender & Eyeglasses \\ \hline
    Pose       & 1.00 & -0.01 & -0.01 &  -0.02 &       0.00 \\ \hline
    Smile      &    - &  1.00 &  0.02 &  -0.08 &      -0.01 \\ \hline
    Age        &    - &     - &  1.00 &   0.42 &       0.35 \\ \hline
    Gender     &    - &     - &     - &   1.00 &       0.47 \\ \hline
    Eyeglasses &    - &     - &     - &      - &       1.00 \\ \hline
  \end{tabular}
  \vspace{-10pt}
\end{table}

Tab.\ref{tab:boundary-correlation} and Tab.\ref{tab:attribute-correlation} report the results.
We can tell that attributes behave similarly under these two metrics, showing that our InterFaceGAN is able to accurately identify the semantics hidden in latent space.
We also find that pose and smile are almost orthogonal to other attributes.
Nevertheless, gender, age, and eyeglasses are highly correlated with each other.
This observation reflects the attribute correlation in the training dataset (\emph{i.e.}, CelebA-HQ \cite{pggan}) to some extent, where male old people are more likely to wear eyeglasses.
This characteristic is also captured by GAN when learning to produce real observation.

\begin{figure*}[t]
  \centering
  \includegraphics[width=0.95\linewidth]{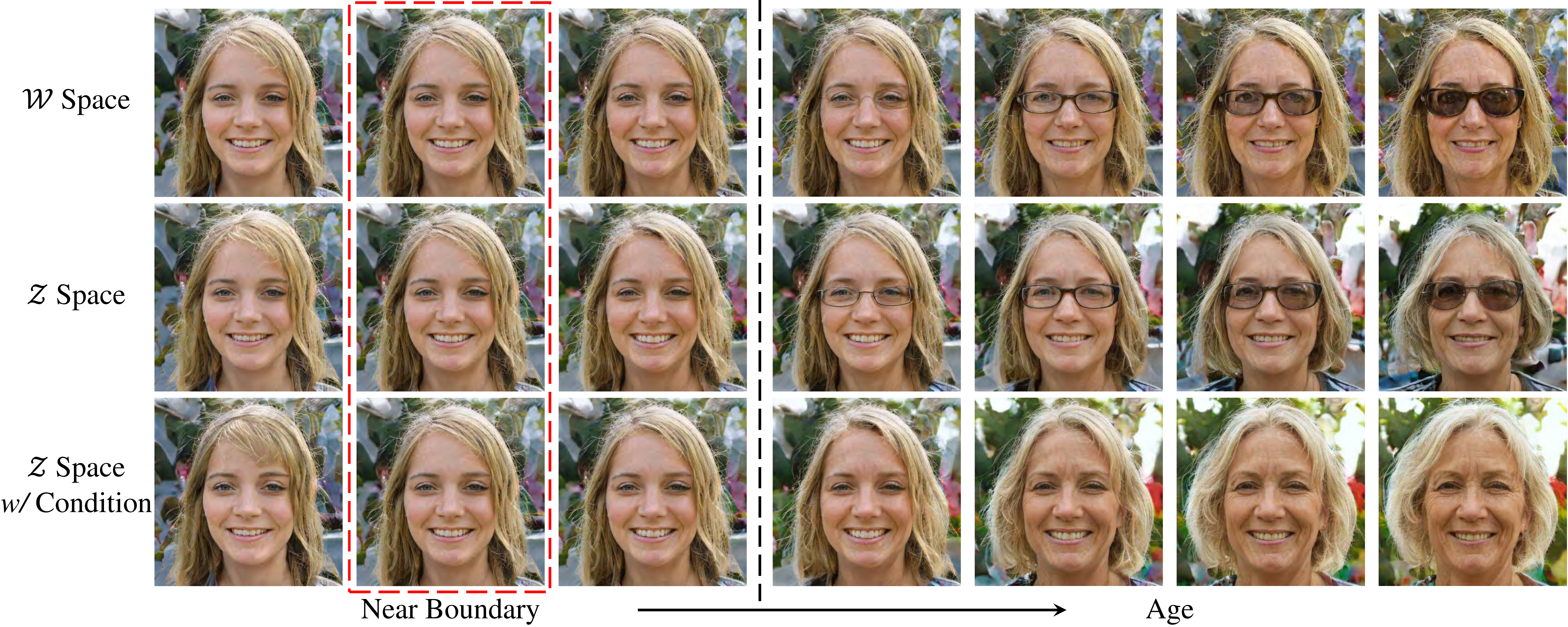}
  \vspace{-10pt}
  \captionsetup{font=small}
  \caption{
    Analysis on the latent space $\Z$ and disentangled latent space $\W$ of StyleGAN \cite{stylegan} by taking age manipulation as an example.
    $\W$ space behaves better for long term manipulation, but the flaw in $\Z$ space can be fixed by projection (\emph{i.e.}, conditional manipulation) to achieve better performance.
  }
  \label{fig:stylegan}
  \vspace{-10pt}
\end{figure*}

\vspace{2pt}\noindent\textbf{Conditional Manipulation.}
To decorrelate different semantics for independent facial attribute editing, we propose conditional manipulation in Sec.\ref{subsec:semantics-manipulation}.
Fig.\ref{fig:condition} shows some results by manipulating one attribute with another one as a condition.
Taking the left sample in Fig.\ref{fig:condition} as an example, the results tend to become male when being edited to get old (first row).
We fix this problem by subtracting its projection onto the gender direction (second row) from age direction, resulting in a new direction.
In this way, we can make sure the gender component is barely affected when the sample is moved along the projected direction (third row).
Fig.\ref{fig:multiple-conditions} shows conditional manipulation with more than one constraint, where we add glasses by conditionally preserving age and gender.
In the beginning, adding eyeglasses is entangled with changing both age and gender.
But we manage to add glasses without affecting age and gender with projection operation.
These two experiments show that our proposed conditional approach helps to achieve independent and precise attribute control.

\subsection{Results on StyleGAN}\label{subsec:results-on-stylegan}
Different from conventional GANs, StyleGAN \cite{stylegan} proposed style-based generator.
Basically, StyleGAN learns to map the latent code from space $\Z$ to another high dimensional space $\W$ before feeding it into the generator.
As pointed out in \cite{stylegan}, $\W$ shows much stronger disentanglement property than $\Z$, since $\W$ is not restricted to any certain distribution and can better model the underlying character of real data.

We did a similar analysis on both $\Z$ and $\W$ spaces of StyleGAN as did to PGGAN and found that $\W$ space indeed learns a more disentangled representation, as pointed out by \cite{stylegan}.
Such disentanglement helps $\W$ space achieve strong superiority over $\Z$ space for attribute editing.
As shown in Fig.\ref{fig:stylegan}, age and eyeglasses are also entangled in StyleGAN model.
Compared to $\Z$ space (second row), $\W$ space (first row) performs better, especially in long-distance manipulation.
Nevertheless, we can use the conditional manipulation trick described in Sec.\ref{subsec:semantics-manipulation} to decorrelate these two attributes in $\Z$ space (third row), resulting in more appealing results.
This trick, however, cannot be applied to $\W$ space.
We found that $\W$ space sometimes captures the attributes correlation that happens in training data and encodes them together as a coupled ``style''.
Taking Fig.\ref{fig:stylegan} as an example, ``age'' and ``eyeglasses'' are supported to be two independent semantics, but StyleGAN actually learns an eyeglasses-included age direction such that this new direction is somehow orthogonal to the eyeglasses direction itself.
In this way, subtracting the projection, which is almost zero, will hardly affect the final results.

\begin{figure*}[t]
  \centering
  \includegraphics[width=0.95\linewidth]{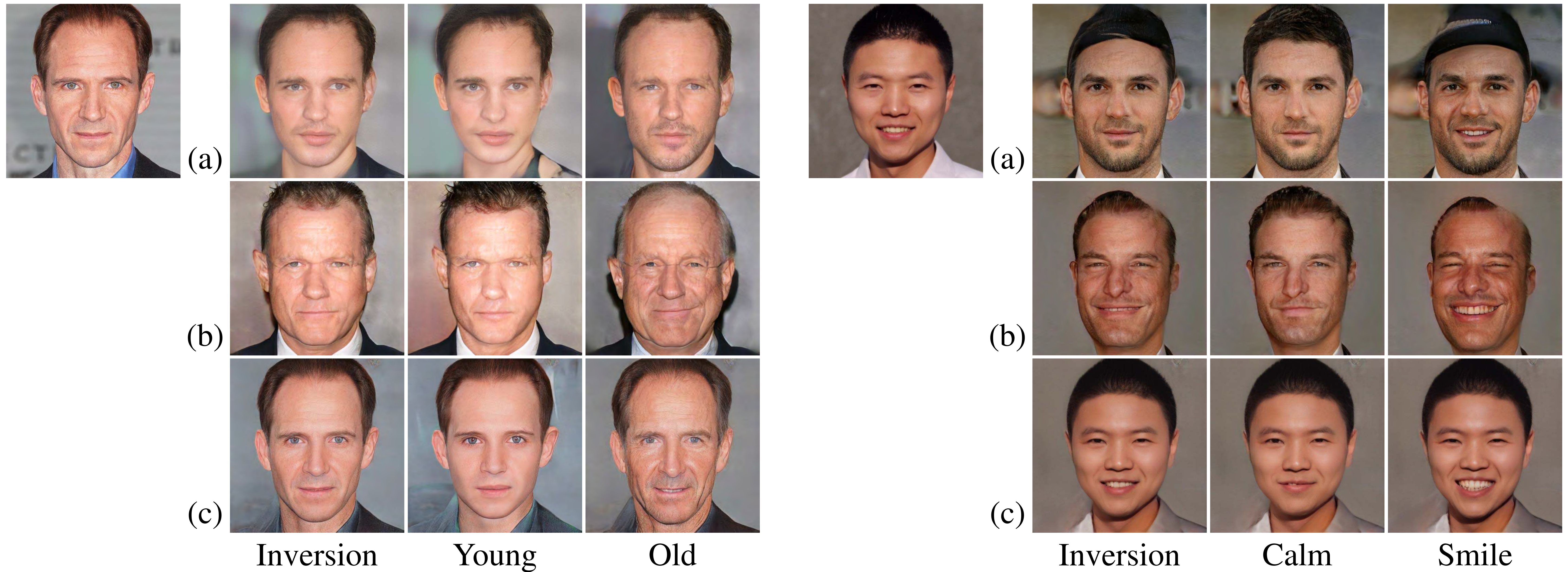}
  \vspace{-10pt}
  \captionsetup{font=small}
  \caption{
    Manipulating real faces with respect to the attributes age and gender, using the pre-trained PGGAN \cite{pggan} and StyleGAN \cite{stylegan}.
    Given an image to edit, we first invert it back to the latent code and then manipulate the latent code with InterFaceGAN.
    On the top left corner is the input real face.
    From top to bottom: (a) PGGAN with optimization-based inversion method, (b) PGGAN with encoder-based inversion method, (c) StyleGAN with optimization-based inversion method.
  }
  \label{fig:real-image-manipulation-inversion}
  \vspace{-10pt}
\end{figure*}

\begin{figure*}[t]
  \centering
  \includegraphics[width=0.9\linewidth]{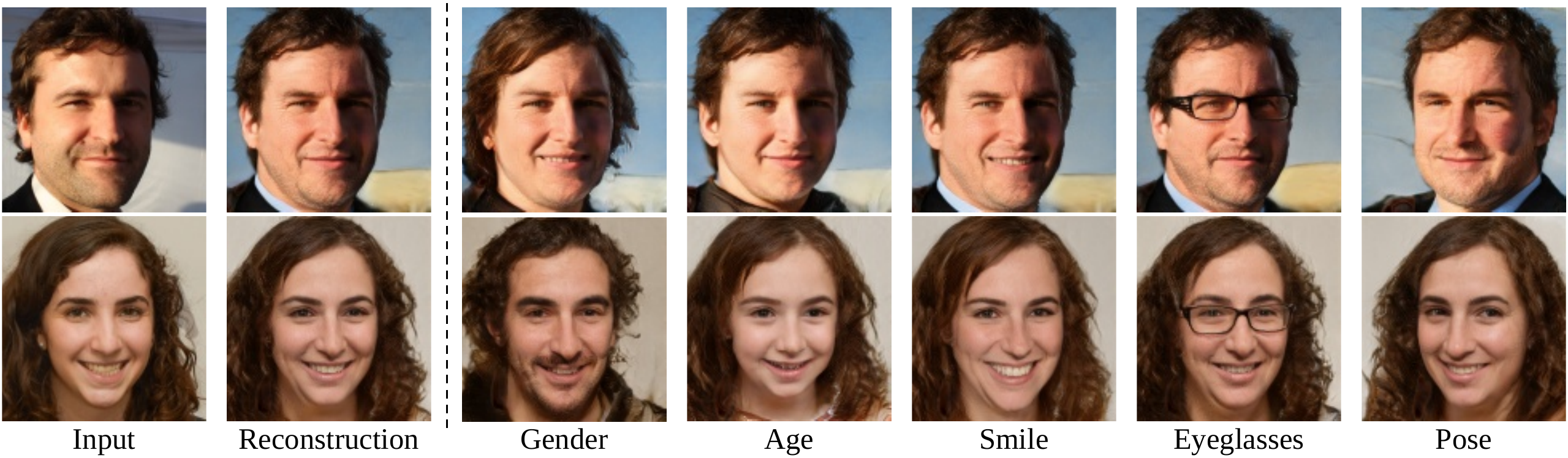}
  \vspace{-10pt}
  \captionsetup{font=small}
  \caption{
    Manipulating real faces with LIA \cite{lia}, which is a encoder-decoder generative model for high-resolution face synthesis.
  }
  \label{fig:real-image-manipulation-encoder}
  \vspace{-13pt}
\end{figure*}

\subsection{Real Image Manipulation}\label{subsec:real-image-manipulation}
In this part, we manipulate real faces with the proposed InterFaceGAN to verify whether the semantic attributes learned by GAN can be applied to real faces.
Recall that InterFaceGAN achieves semantic face editing by moving the latent code along a certain direction.
Accordingly, we need to first invert the given real image back to the latent code.
It turns out to be a non-trivial task because GANs do not fully capture all the modes as well as the diversity of the true distribution.
To invert a pre-trained GAN model, there are two typical approaches.
One is the optimization-based approach, which directly optimizes the latent code with the fixed generator to minimize the pixel-wise reconstruction error \cite{invertibility}.
The other is the encoder-based, where an extra encoder network is trained to learn the inverse mapping \cite{zhu2016generative}.
We tested the two baseline approaches on PGGAN and StyleGAN.

Results are shown in Fig.\ref{fig:real-image-manipulation-inversion}.
We can tell that both optimization-based (first row) and encoder-based (second row) methods show poor performance when inverting PGGAN.
This can be imputed to the strong discrepancy between training and testing data distributions.
For example, the model tends to generate Western people even the input is an Easterner (see the right example in Fig.\ref{fig:real-image-manipulation-inversion}).
Even unlike the inputs, however, the inverted images can still be semantically edited with InterFaceGAN.
Compared to PGGAN, the results on StyleGAN (third row) are much better.
Here, we treat the layer-wise styles (\emph{i.e.}, $\w$ for all layers) as the optimization target.
When editing an instance, we push all style codes towards the same direction.
As shown in Fig.\ref{fig:real-image-manipulation-inversion}, we successfully change the attributes of real face images \emph{without} retraining StyleGAN but leveraging the interpreted semantics from latent space.

We also test InterFaceGAN on encoder-decoder generative models, which train an encoder together with the generator and discriminator.
After the model converges, the encoder can be directly used for inference to map a given image to latent space.
We apply our method to interpret the latent space of the recent encoder-decoder model LIA \cite{lia}.
The manipulation result is shown in Fig.\ref{fig:real-image-manipulation-encoder} where we successfully edit the input faces with various attributes, like age and face pose.
It suggests that the latent code in the encoder-decoder based generative models also supports semantic manipulation.
In addition, compared to Fig.\ref{fig:real-image-manipulation-inversion} (b) where the encoder is separately learned after the GAN model is well-prepared, the encoder trained together with the generator gives better reconstruction as well as manipulation results.

\section{Conclusion}\label{sec:conclusion}
We propose InterFaceGAN to interpret the semantics encoded in the latent space of GANs.
By leveraging the interpreted semantics as well as the proposed conditional manipulation technique, we are able to precisely control the facial attributes with any fixed GAN model, even turning unconditional GANs to controllable GANs.
Extensive experiments suggest that InterFaceGAN can also be applied to real image editing.

\vspace{2pt}\noindent\textbf{Acknowledgement:}
This work is supported in part by the Early Career Scheme (ECS) through the Research Grants Council of Hong Kong under Grant No.24206219 and in part by SenseTime Collaborative Grant.

\appendix
\section*{Appendix}

\section{Overview}\label{appendix:overview}
This appendix contains the following information:
\begin{itemize}
  \item We introduce the implementation details of the proposed InterFaceGAN in Sec.\ref{appendix:implementation-details}.
  \item We provide the detailed proof of \emph{Property 2} in the main paper in Sec.\ref{appendix:proof}.
  \item Please also refer to \href{https://www.youtube.com/watch?v=uoftpl3Bj6w}{this video} to see continuous attribute editing results.
\end{itemize}

\section{Implementation Details}\label{appendix:implementation-details}
We choose five key facial attributes for analysis, including pose, smile (expression), age, gender, and eyeglasses.
The corresponding positive directions are defined as turning right, laughing, getting old, changing to male, and wearing eyeglasses. Note that we can always plug in more attributes easily as long as the attribute detector is available.

To better predict these attributes from synthesized images, we train an auxiliary attribute prediction model using the annotations from the CelebA dataset \cite{celeba} with ResNet-50 network \cite{resnet}.
This model is trained with multi-task losses to simultaneously predict smile, age, gender, eyeglasses, as well as the 5-point facial landmarks.
Here, the facial landmarks will be used to compute yaw pose, which is also treated as a binary attribute (left or right) in further analysis.
Besides the landmarks, all other attributes are learned as bi-classification problem with softmax cross-entropy loss, while landmarks are optimized with $l_2$ regression loss.
As images produced by PGGAN and StyleGAN are with $1024\times1024$ resolution, we resize them to $224\times224$ before feeding them to the attribute model.

Given the pre-trained GAN model, we synthesize $500K$ images by randomly sampling the latent space.
There are mainly two reasons in preparing such large-scale data:
(i) to eliminate the randomness caused by sampling and make sure the distribution of the latent codes is as expected,
and (ii) to get enough wearing-glasses samples, which are really rare in PGGAN model.

To find the semantic boundaries in the latent space, we use the pre-trained attribute prediction model to assign attribute scores for all $500K$ synthesized images.
For each attribute, we sort the corresponding scores, and choose $10K$ samples with highest scores and $10K$ with lowest ones as candidates.
The reason in doing so is that the prediction model is not absolutely accurate and may produce wrong prediction for ambiguous samples, \emph{e.g.}, middle-aged person for age attribute.
We then randomly choose 70\% samples from the candidates as the training set to learn a linear SVM, resulting in a decision boundary.
Recall that, normal directions of all boundaries are normalized to unit vectors.
Remaining 30\% are used for verifying how the linear classifier behaves.
Here, for SVM training, the inputs are the $512d$ latent codes, while the binary labels are assigned by the auxiliary attribute prediction model.

\section{Proof}\label{appendix:proof}
In this part, we provide detailed proof of \emph{Property 2} in the main paper.
Recall this property as follow.

\vspace{2pt}\noindent\emph{\textbf{Property 2}
Given $\n\in\R^d$ with $\n^T\n=1$, which defines a hyperplane, and a multivariate random variable $\z\sim\Norm(\0,\I_d)$, we have $\Prob(|\n^T\z|\leq2\alpha~\sqrt{\frac{d}{d-2}})\geq(1-3e^{-c d})(1-\frac{2}{\alpha}e^{-\alpha^2/2})$ for any $\alpha\geq1$ and $d\geq4$. Here $\Prob(\cdot)$ stands for probability and $c$ is a fixed positive constant.
}

\vspace{2pt}\noindent\emph{Proof.}

Without loss of generality, we fix $\n$ to be the first coordinate vector.
Accordingly, it suffices to prove that $\Prob(|z_1|\leq2\alpha~\sqrt{\frac{d}{d-2}})\geq(1-3e^{-c d})(1-\frac{2}{\alpha}e^{-\alpha^2/2})$, where $z_1$ denotes the first entry of $\z$.

\begin{figure}
  \centering
  \includegraphics[width=0.9\linewidth]{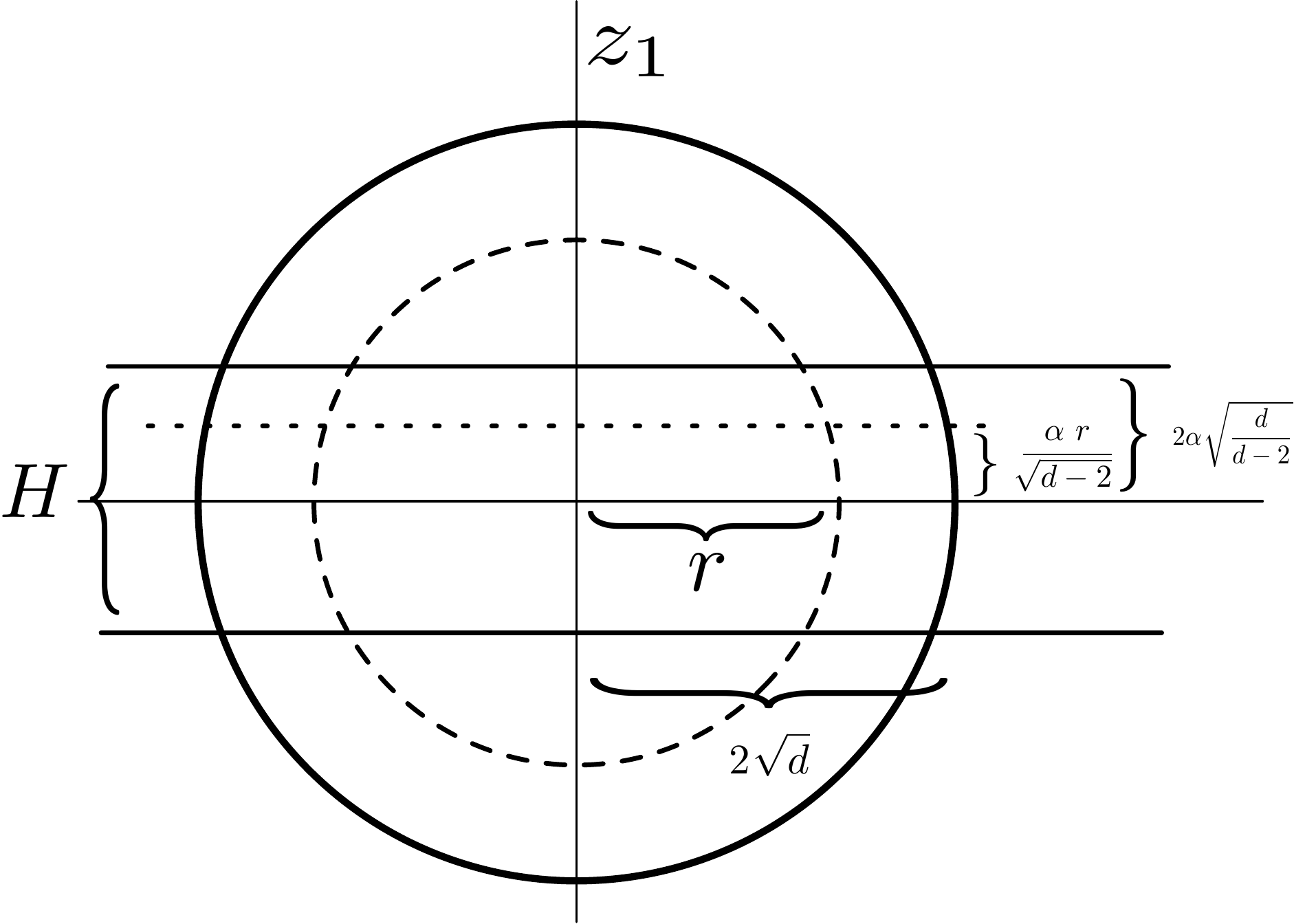}
  \vspace{-5pt}
  \captionsetup{font=small}
  \caption{
    Illustration of \emph{Property 2}, which shows that most of the probability mass of high-dimensional Gaussian distribution lies in the thin slab near the ``equator''.
  }
  \vspace{-10pt}
  \label{fig:property_2}
\end{figure}

As shown in Fig.\ref{fig:property_2}, let $H$ denote the set
\begin{align}
\{\z\sim\N(\0,\I_d):||\z||_2\leq2\sqrt{d}, |z_1|\leq2\alpha\sqrt{\frac{d}{d-2}}\}, \nonumber
\end{align}
where $||\cdot||_2$ stands for the $l_2$ norm.
Obviously, we have $\Prob(H)\leq\Prob(|z_1|\leq2\alpha\sqrt{\frac{d}{d-2}})$.
Now, we will show $\Prob(H)\geq(1-3e^{-c d})(1-\frac{2}{\alpha}e^{-\alpha^2/2})$

Considering the random variable $R=||\z||_2$, with cumulative distribution function $F(R\leq r)$ and density function $f(r)$, we have
\begin{align}
  \Prob(H) &= \Prob(|z_1|\leq2\alpha\sqrt{\frac{d}{d-2}}|R\leq2\sqrt{d})\Prob(R\leq2\sqrt{d}) \nonumber \\
           &= \int_0^{2\sqrt{d}}\Prob(|z_1|\leq2\alpha\sqrt{\frac{d}{d-2}}|R=r)f(r)dr. \nonumber
\end{align}

According to \emph{Theorem 1} below, when $r\leq2\sqrt{d}$, we have
\begin{align}
   \Prob(H) &= \int_0^{2\sqrt{d}}\Prob(|z_1|\leq2\alpha\sqrt{\frac{d}{d-2}}|R=r)f(r)dr \nonumber \\
            &= \int_0^{2\sqrt{d}}\Prob(|z_1|\leq\frac{2\sqrt{d}}{r}\frac{\alpha}{\sqrt{d-2}}|R=1)f(r)dr \nonumber \\
            &\geq \int_0^{2\sqrt{d}}\Prob(|z_1|\leq\frac{\alpha}{\sqrt{d-2}}|R=1)f(r)dr \nonumber \\
            &\geq \int_0^{2\sqrt{d}}(1-\frac{2}{\alpha}e^{-\alpha^2/2})f(r)dr \nonumber \\
            &= (1-\frac{2}{\alpha}e^{-\alpha^2/2})\int_0^{2\sqrt{d}}f(r)dr \nonumber \\
            &= (1-\frac{2}{\alpha}e^{-\alpha^2/2})\Prob(0\leq R\leq2\sqrt{d}). \nonumber
\end{align}

Then, according to \emph{Theorem 2} below, by setting $\beta=\sqrt{d}$, we have
\begin{align}
  \Prob(H) &= (1-\frac{2}{\alpha}e^{-\alpha^2/2})\Prob(0\leq R\leq2\sqrt{d}) \nonumber \\
           &\geq (1-\frac{2}{\alpha}e^{-\alpha^2/2})(1-3e^{-c d}). \nonumber
\end{align}

Q.E.D.

\vspace{2pt}\noindent\emph{\textbf{Theorem 1}
Given a unit spherical $\{\z\in\R^d:||\z||_2=1\}$, we have $\Prob(|z_1|\leq\frac{\alpha}{\sqrt{d-2}})\geq1-\frac{2}{\alpha}e^{-\alpha^2/2}$ for any $\alpha\geq1$ and $d\geq4$.}

\vspace{2pt}\noindent\emph{Proof.}

By symmetry, we just prove the case where $z_1\geq0$.
Also, we only consider about the case where $\frac{\alpha}{\sqrt{d-2}} \leq 1$.

\begin{figure}
  \centering
  \includegraphics[width=0.9\linewidth]{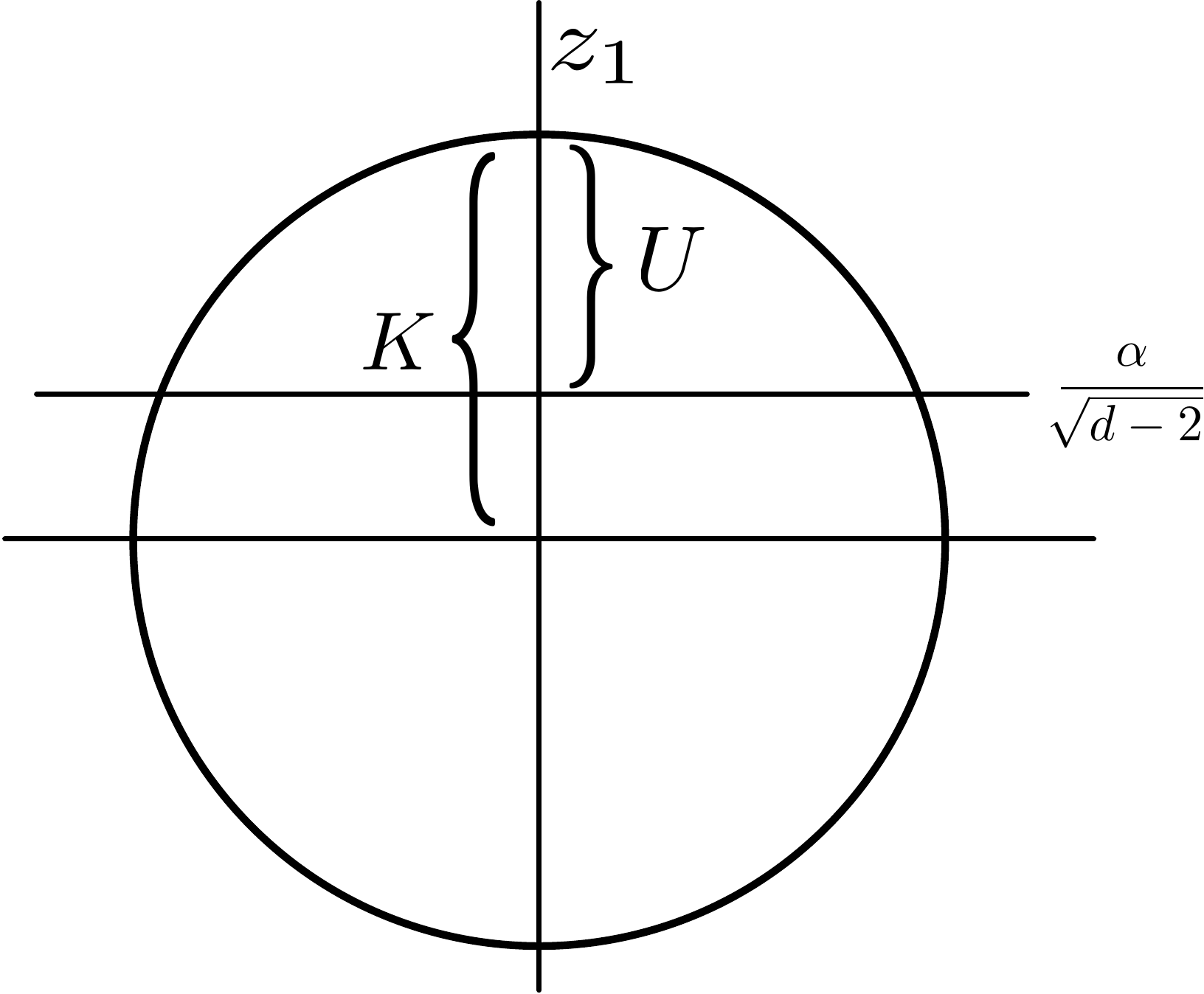}
  \vspace{-5pt}
  \caption{
    Diagram for \emph{Theorem 1}.
  }
  \label{fig:theorem_1}
  \vspace{-10pt}
\end{figure}

Let $U$ denote the set $\{\z\in\R^d:||\z||_2=1,z_1\geq\frac{\alpha}{\sqrt{d-2}}\}$, and $K$ denote the set $\{\z\in\R^d:||\z||_2=1,z_1\geq0\}$.
It suffices to prove that the surface of $U$ area and the surface of $K$ area in Fig.\ref{fig:theorem_1} satisfy
\begin{align}
  \frac{surf(U)}{surf(K)}\leq\frac{2}{\alpha}e^{-\alpha^2/2}, \nonumber
\end{align}
where $surf(\cdot)$ stands for the surface area of a high dimensional geometry.
Let $A(d)$ denote the surface area of a $d$-dimensional unit-radius ball. Then, we have
\begin{align}
  surf(U) &=\int_{\frac{\alpha}{\sqrt{d-2}}}^{1} (1-z_1^2)^{\frac{d-2}{2}}A(d-1)dz_1 \nonumber \\
          &\leq \int_{\frac{\alpha}{\sqrt{d-2}}}^{1} e^{-\frac{d-2}{2}z_1^2}A(d-1)dz_1 \nonumber \\
          &\leq \int_{\frac{\alpha}{\sqrt{d-2}}}^{1} \frac{z_1\sqrt{d-2}}{\alpha}e^{-\frac{d-2}{2}z_1^2}A(d-1)dz_1 \nonumber \\
          &\leq \int_{\frac{\alpha}{\sqrt{d-2}}}^{\infty} \frac{z_1\sqrt{d-2}}{\alpha}e^{-\frac{d-2}{2}z_1^2}A(d-1)dz_1 \nonumber \\
          &= \frac{A(d-1)}{\alpha\sqrt{d-2}}e^{-\alpha^2/2}. \nonumber
\end{align}

Similarly, we have
\begin{align}
  surf(K) &=\int_{0}^{1} (1-z_1^2)^{\frac{d-2}{2}}A(d-1)dz_1 \nonumber \\
          &\geq \int_{0}^{\frac{1}{\sqrt{d-2}}} (1-z_1^2)^{\frac{d-2}{2}}A(d-1)dz_1 \nonumber \\
          &\geq \frac{1}{\sqrt{d-2}} (1-\frac{1}{d-2})^{\frac{d-2}{2}}A(d-1). \nonumber
\end{align}

Considering the fact that $(1-x)^a\geq1-ax$ for any $a\geq1$ and $0\leq x\leq1$, we have
\begin{align}
  surf(K) &\geq \frac{1}{\sqrt{d-2}} (1-\frac{1}{d-2})^{\frac{d-2}{2}}A(d-1) \nonumber \\
          &\geq \frac{1}{\sqrt{d-2}} (1-\frac{1}{d-2}\frac{d-2}{2})A(d-1) \nonumber \\
          &= \frac{A(d-1)}{2\sqrt{d-2}}. \nonumber
\end{align}

Accordingly,
\begin{align}
  \frac{surf(U)}{surf(K)} \leq \frac{\frac{A(d-1)}{\alpha\sqrt{d-2}}e^{-\alpha^2/2}}{\frac{A(d-1)}{2\sqrt{d-2}}}
                           = \frac{2}{\alpha}e^{-\alpha^2/2}. \nonumber
\end{align}

Q.E.D.

\vspace{2pt}\noindent\emph{\textbf{Theorem 2 (Gaussian Annulus Theorem  \cite{foundations_of_data_science})}
For a $d$-dimensional spherical Gaussian with unit variance in each direction, for any $\beta\leq\sqrt{d}$, all but at most $3e^{-c\beta^2}$ of the probability mass lies within the annulus $\sqrt{d}-\beta\leq||\z||_2\leq\sqrt{d}+\beta$, where $c$ is a fixed positive constant.
}

That is to say, given $\z\sim\N(\0, \I_d)$, $\beta\leq\sqrt{d}$, and a constant $c>0$, we have
\begin{align}
  \Prob(\sqrt{d}-\beta\leq||\z||_2\leq\sqrt{d}+\beta)\geq(1-3e^{-c\beta^2}). \nonumber
\end{align}

{\small
\bibliographystyle{ieee_fullname}
\bibliography{ref}

\begin{thebibliography}{10}\itemsep=-1pt

\bibitem{wgan}
Martin Arjovsky, Soumith Chintala, and L{\'e}on Bottou.
\newblock Wasserstein generative adversarial networks.
\newblock In {\em ICML}, 2017.

\bibitem{latent_oddity}
Georgios Arvanitidis, Lars~Kai Hansen, and S{\o}ren Hauberg.
\newblock Latent space oddity: on the curvature of deep generative models.
\newblock In {\em ICLR}, 2018.

\bibitem{opensetgan}
Jianmin Bao, Dong Chen, Fang Wen, Houqiang Li, and Gang Hua.
\newblock Towards open-set identity preserving face synthesis.
\newblock In {\em CVPR}, 2018.

\bibitem{gan_dissection}
David Bau, Jun-Yan Zhu, Hendrik Strobelt, Bolei Zhou, Joshua~B. Tenenbaum,
  William~T. Freeman, and Antonio Torralba.
\newblock Visualizing and understanding generative adversarial networks.
\newblock In {\em ICLR}, 2019.

\bibitem{bau2019seeing}
David Bau, Jun-Yan Zhu, Jonas Wulff, William Peebles, Hendrik Strobelt, Bolei
  Zhou, and Antonio Torralba.
\newblock Seeing what a gan cannot generate.
\newblock In {\em ICCV}, 2019.

\bibitem{began}
David Berthelot, Thomas Schumm, and Luke Metz.
\newblock Began: Boundary equilibrium generative adversarial networks.
\newblock {\em arXiv preprint arXiv:1703.10717}, 2017.

\bibitem{glo}
Piotr Bojanowski, Armand Joulin, David Lopez-Pas, and Arthur Szlam.
\newblock Optimizing the latent space of generative networks.
\newblock In {\em ICML}, 2018.

\bibitem{biggan}
Andrew Brock, Jeff Donahue, and Karen Simonyan.
\newblock Large scale {GAN} training for high fidelity natural image synthesis.
\newblock In {\em ICLR}, 2019.

\bibitem{gan_metrics}
Nutan Chen, Alexej Klushyn, Richard Kurle, Xueyan Jiang, Justin Bayer, and
  Patrick van~der Smagt.
\newblock Metrics for deep generative models.
\newblock In {\em AISTAT}, 2018.

\bibitem{infogan}
Xi Chen, Yan Duan, Rein Houthooft, John Schulman, Ilya Sutskever, and Pieter
  Abbeel.
\newblock Infogan: Interpretable representation learning by information
  maximizing generative adversarial nets.
\newblock In {\em NeurIPS}, 2016.

\bibitem{sdgan}
Chris Donahue, Akshay Balsubramani, Julian McAuley, and Zachary~C. Lipton.
\newblock Semantically decomposing the latent spaces of generative adversarial
  networks.
\newblock In {\em ICLR}, 2018.

\bibitem{bigan}
Jeff Donahue, Philipp Kr{\"a}henb{\"u}hl, and Trevor Darrell.
\newblock Adversarial feature learning.
\newblock In {\em ICLR}, 2017.

\bibitem{ali}
Vincent Dumoulin, Ishmael Belghazi, Ben Poole, Olivier Mastropietro, Alex Lamb,
  Martin Arjovsky, and Aaron Courville.
\newblock Adversarially learned inference.
\newblock In {\em ICLR}, 2017.

\bibitem{goetschalckx2019ganalyze}
Lore Goetschalckx, Alex Andonian, Aude Oliva, and Phillip Isola.
\newblock Ganalyze: Toward visual definitions of cognitive image properties.
\newblock In {\em ICCV}, 2019.

\bibitem{gan}
Ian Goodfellow, Jean Pouget-Abadie, Mehdi Mirza, Bing Xu, David Warde-Farley,
  Sherjil Ozair, Aaron Courville, and Yoshua Bengio.
\newblock Generative adversarial nets.
\newblock In {\em NeurIPS}, 2014.

\bibitem{gu2020image}
Jinjin Gu, Yujun Shen, and Bolei Zhou.
\newblock Image processing using multi-code gan prior.
\newblock In {\em CVPR}, 2020.

\bibitem{wgan_gp}
Ishaan Gulrajani, Faruk Ahmed, Martin Arjovsky, Vincent Dumoulin, and Aaron~C
  Courville.
\newblock Improved training of wasserstein gans.
\newblock In {\em NeurIPS}, 2017.

\bibitem{resnet}
Kaiming He, Xiangyu Zhang, Shaoqing Ren, and Jian Sun.
\newblock Deep residual learning for image recognition.
\newblock In {\em CVPR}, 2016.

\bibitem{foundations_of_data_science}
John Hopcroft and Ravi Kannan.
\newblock {\em Foundations of Data Science}.
\newblock 2014.

\bibitem{gansteerability}
Ali Jahanian, Lucy Chai, and Phillip Isola.
\newblock On the "steerability" of generative adversarial networks.
\newblock In {\em ICLR}, 2020.

\bibitem{pggan}
Tero Karras, Timo Aila, Samuli Laine, and Jaakko Lehtinen.
\newblock Progressive growing of {GAN}s for improved quality, stability, and
  variation.
\newblock In {\em ICLR}, 2018.

\bibitem{stylegan}
Tero Karras, Samuli Laine, and Timo Aila.
\newblock A style-based generator architecture for generative adversarial
  networks.
\newblock In {\em CVPR}, 2019.

\bibitem{kuhnel2018latent}
Line Kuhnel, Tom Fletcher, Sarang Joshi, and Stefan Sommer.
\newblock Latent space non-linear statistics.
\newblock {\em arXiv preprint arXiv:1805.07632}, 2018.

\bibitem{feature_based_metrics}
Samuli Laine.
\newblock Feature-based metrics for exploring the latent space of generative
  models.
\newblock In {\em ICLR Workshop}, 2018.

\bibitem{fader}
Guillaume Lample, Neil Zeghidour, Nicolas Usunier, Antoine Bordes, Ludovic
  Denoyer, and Marc'Aurelio Ranzato.
\newblock Fader networks: Manipulating images by sliding attributes.
\newblock In {\em NeurIPS}, 2017.

\bibitem{celeba}
Ziwei Liu, Ping Luo, Xiaogang Wang, and Xiaoou Tang.
\newblock Deep learning face attributes in the wild.
\newblock In {\em ICCV}, 2015.

\bibitem{invertibility}
Fangchang Ma, Ulas Ayaz, and Sertac Karaman.
\newblock Invertibility of convolutional generative networks from partial
  measurements.
\newblock In {\em NeurIPS}, 2018.

\bibitem{sngan}
Takeru Miyato, Toshiki Kataoka, Masanori Koyama, and Yuichi Yoshida.
\newblock Spectral normalization for generative adversarial networks.
\newblock In {\em ICLR}, 2018.

\bibitem{acgan}
Augustus Odena, Christopher Olah, and Jonathon Shlens.
\newblock Conditional image synthesis with auxiliary classifier gans.
\newblock In {\em ICML}, 2017.

\bibitem{icgan}
Guim Perarnau, Joost Van De~Weijer, Bogdan Raducanu, and Jose~M {\'A}lvarez.
\newblock Invertible conditional gans for image editing.
\newblock In {\em NeurIPS Workshop}, 2016.

\bibitem{dcgan}
Alec Radford, Luke Metz, and Soumith Chintala.
\newblock Unsupervised representation learning with deep convolutional
  generative adversarial networks.
\newblock In {\em ICLR}, 2016.

\bibitem{riemannian_geometry}
Hang Shao, Abhishek Kumar, and P Thomas~Fletcher.
\newblock The riemannian geometry of deep generative models.
\newblock In {\em CVPR Workshop}, 2018.

\bibitem{faceidgan}
Yujun Shen, Ping Luo, Junjie Yan, Xiaogang Wang, and Xiaoou Tang.
\newblock Faceid-gan: Learning a symmetry three-player gan for
  identity-preserving face synthesis.
\newblock In {\em CVPR}, 2018.

\bibitem{facefeatgan}
Yujun Shen, Bolei Zhou, Ping Luo, and Xiaoou Tang.
\newblock Facefeat-gan: a two-stage approach for identity-preserving face
  synthesis.
\newblock {\em arXiv preprint arXiv:1812.01288}, 2018.

\bibitem{drgan}
Luan Tran, Xi Yin, and Xiaoming Liu.
\newblock Disentangled representation learning gan for pose-invariant face
  recognition.
\newblock In {\em CVPR}, 2017.

\bibitem{feature_interpolation}
Paul Upchurch, Jacob Gardner, Geoff Pleiss, Robert Pless, Noah Snavely, Kavita
  Bala, and Kilian Weinberger.
\newblock Deep feature interpolation for image content changes.
\newblock In {\em CVPR}, 2017.

\bibitem{elegant}
Taihong Xiao, Jiapeng Hong, and Jinwen Ma.
\newblock Elegant: Exchanging latent encodings with gan for transferring
  multiple face attributes.
\newblock In {\em ECCV}, 2018.

\bibitem{yang2019semantic}
Ceyuan Yang, Yujun Shen, and Bolei Zhou.
\newblock Semantic hierarchy emerges in deep generative representations for
  scene synthesis.
\newblock {\em arXiv preprint arXiv:1911.09267}, 2019.

\bibitem{ffgan}
Xi Yin, Xiang Yu, Kihyuk Sohn, Xiaoming Liu, and Manmohan Chandraker.
\newblock Towards large-pose face frontalization in the wild.
\newblock In {\em ICCV}, 2017.

\bibitem{sagan}
Han Zhang, Ian Goodfellow, Dimitris Metaxas, and Augustus Odena.
\newblock Self-attention generative adversarial networks.
\newblock In {\em ICML}, 2019.

\bibitem{lia}
Jiapeng Zhu, Deli Zhao, and Bo Zhang.
\newblock Lia: Latently invertible autoencoder with adversarial learning.
\newblock {\em arXiv preprint arXiv:1906.08090}, 2019.

\bibitem{zhu2016generative}
Jun-Yan Zhu, Philipp Kr{\"a}henb{\"u}hl, Eli Shechtman, and Alexei~A Efros.
\newblock Generative visual manipulation on the natural image manifold.
\newblock In {\em ECCV}, 2016.

\end{thebibliography}
}

\end{document}